# L1-2D²PCANet: A Deep Learning Network for Face Recognition

Yun-Kun Li[1], Xiao-Jun Wu[1*], Josef Kittler[2]

[1] Jiangsu Provincial Engineering Laboratory of Pattern Recognition and Computational Intelligence,
Jiangnan University, 214122, Wuxi, China

[2] Center for Vision, Speech and Signal Processing(CVSSP), University of Surry, GU2 7XH, Guildford, UK

{m15979095336, xiaojun_wu_jnu}@163.com, j.kittler@surrey.ac.uk

*Abstract*—In this paper, we propose a novel deep learning network L1-2D²PCANet for face recognition, which is based on L1-norm-based two-directional two-dimensional principal component analysis (L1-2D²PCA). In our network, the role of L1-2D²PCA is to learn the filters of multiple convolution layers. After the convolution layers, we deploy binary hashing and block-wise histogram for pooling. We test our network on some benchmark facial datasets YALE, AR, Extended Yale B, LFW-a and FERET with CNN, PCANet, 2DPCANet and L1-PCANet as comparison. The results show that the recognition performance of L1-2D²PCANet in all tests is better than baseline networks, especially when there are outliers in the test data. Owing to the L1-norm, L1-2D²PCANet is robust to outliers and changes of the training images.

*Keywords—face recognition, deep learning, L1-2D²PCA, outlier*

## I. Introduction

In pattern recognition and computer vision, face recognition is a very important research field [1,2,3,4]. Due to the complexity of facial features and the difficulty of manual feature selection [1], it is commonly agreed that the best features can be obtained by using unsupervised feature extraction methods [3,4].

Recently, with Google AlphaGo Zero defeating many Go masters, deep learning has received intensive attentions [5,6]. As a classical deep learning model, Convolution Neural Networks (CNNs) with convolution and pooling layers have achieved astonishing results in many image recognition tasks, reaching an unprecedented accuracy [7,8]. However, CNN still has many shortcomings. During training a CNN model, researchers need to obtain a huge amount of parameters, which leads to large computational cost.

To solve this problem, researchers are committed to find a simple CNN model which requires a small number of parameters. Chan et al. proposed PCANet [9], which is a simple deep learning network based on unsupervised learning. PCANet uses PCA to learn the filters and deploys simple binary hashing and block histograms for indexing and pooling. Unlike other CNNs that learn filters by back propagation, PCANet learns filters using the PCA method. Thus PCANet requires less computational cost, less time and storage space. The experimental results show the astonishing performance of PCANet.

The PCA method used by PCANet is based on 1D vectors. Before deploying PCA, we need to convert 2D image matrices into 1D vectors which will cause two major problems: (1) Some spatial information of image is implied in the 2D structure of the image. Obviously, the intrinsic information is discarded when the image matrix is converted into 1D vector. (2) the long 1D vector leads to the requirement of large computational time and storage space in computing the eigenvectors. To solve these problems, Yu et al. proposed two-dimensional principal component analysis network (2DPCANet) [10], which replaces PCA with 2DPCA [11,12,13].

However, Both PCA and 2DPCA are based on L2-norm method. It is well known that the methods based on L2-norm are sensitive to outliers so that data with outliers can totally ruin the results from the desired methods. To solve this problem, Nojun et al. proposed a novel PCA method based on L1-norm [14]. L1-norm is widely considered to be more robust to outliers [15,16]. L1-PCA adopts the L1-norm for measuring the reconstruction error. On this basis, Li et al. proposed L1-norm-based 2DPCA [17].

In this paper, L1-norm was introduced into PCANet to get L1-PCANet. Then we generalize L1-PCANet to L1-2D²PCANet. L1-2D²PCANet and 2DPCANet share the same structure to generate the feature of input data but L1-2D²PCANet learns filters by L1-2DPCA. In addition, we use Support Vector Machine (SVM) as classifiers for the features generated by the networks. To test the performance of L1-2D²PCANet, we compare it with other three networks (PCANet, 2DPCANet and L1-PCANet) on Yale, AR [18], Extended Yale B [19], LFW-a [20] and FERET [21] face databases.

The rest of paper is organized as follows. Section II reviews related work on L1-PCA and L1-2DPCA. L1-PCANet is given in Section III and L1-2D²PCANet is given in Section IV. Section V reports the results and analysis of the experiments and Section VI concludes this paper.

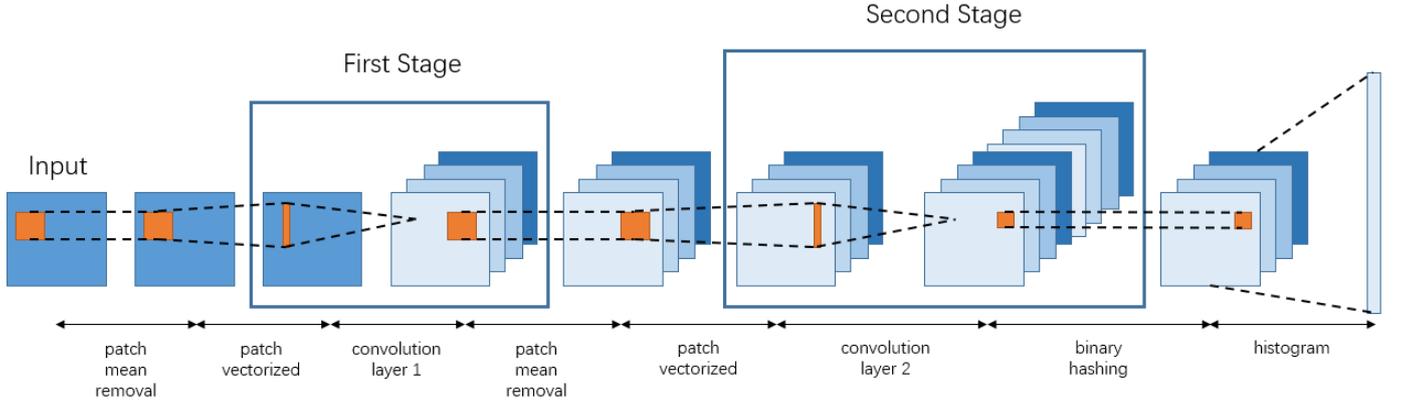

Fig.1 The illustration of two-layer L1-PCANet

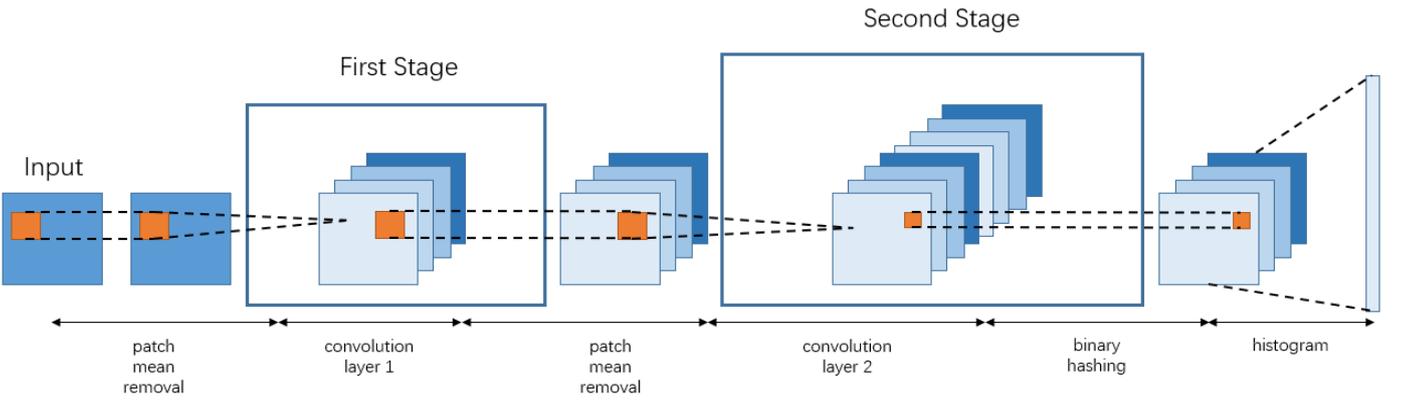

Fig.2 The illustration of two-layer L1-2D$^2$PCANet

## II. RELATERD WORK

### A. *L1-norm-based PCA*

The proposed L1-PCANet is based on L1-PCA [14,15]. PCA-L1 is considered as the simplest and most efficient among many models of L1-norm-based PCA. Let $X = [x_1, x_2, \ldots, x_N] \in \mathbb{R}^{D \times N}$, with $x_i = mat_D(I_i) \in \mathbb{R}^{D \times 1}$ ($i = 1, 2, \ldots, N$). The $mat_D(I)$ is a function that maps a matrix $I \in \mathbb{R}^{m \times n}$ to a vector $v \in \mathbb{R}^{D \times 1}$ and $D = m \times n$. Suppose $w \in \mathbb{R}^{D \times 1}$ be the principal vector to be obtained. Here, we set the number of principal vectors to one to simplify the procedure. The objective of PCA-L1 is to maximize the L1-norm variance in the feature space as follows:

$$f(w) = \|w^T X\|_1 = \sum_{i=1}^{N} |w^T x_i|$$

$$subject\ to\ \|w\|_2 = 1, \qquad (1)$$

where $\|\cdot\|$ denotes L2-norm and $|\cdot|$ denotes L1-norm.

To solve the computational problems posed by the symbol of absolute value, we introduce a polarity parameter $p_i$ in Equation (1)

$$p_i = \begin{cases} 1, & when\ w^T x_i \geq 0 \\ -1, & when\ w^T x_i < 0. \end{cases} \qquad (2)$$

By introducing $p_i$, Equation 1 can be rewritten as:

$$f(w) = \sum_{i=1}^{N} p_i w^T x_i. \qquad (3)$$

The process of maximization is achieved by Algorithm 1.

Here, $t$ denotes the number of iterations and $w(t)$ and $p_i(t)$ denote $w$ and $p_i$ during iteration $t$.

By the above algorithm, we can obtain the first principal vector $w_1^*$. To compute $w_k^*(k > 1)$, we have to update the training data as:

$$x_i^k = x_i^{k-1} - x_i^{k-1}(w_{k-1}^* w_{k-1}^{*T}). \qquad (4)$$

| **Algorithm 1**: L1-PCA method |
|---|

Input:
- training set: $X = [x_1, x_2, ..., x_N] \in \mathbb{R}^{D \times N}$

Output:
- filters $w^*$

1: set $w(0) = 0$ and $t = 0$

2: For all $i \in \{1, 2, ..., N\}$, calculate $p_i(t)$ by using Equation (2)

3: Let $t = t + 1$ and $w(t) = \sum_{i=1}^{N} p_i(t-1)x_i$. Then let $w(t) = w(t)/\|w(t)\|_2$

4: If $w(t) \neq w(t-1)$, go back to Step 2. Otherwise, set $w^* = w(t)$ and stop.

### B. L1-norm-based 2DPCA

In this section, we extend PCA-L1 to L1-2DPCA [17]. As mentioned above, 2DPCA compute eigenvectors with 2D input. Suppose $I_i(i = 1,2,...,N)$ denote $N$ input training images and $D = m \times n$ being the image size. Let $w \in \mathbb{R}^{w \times 1}$ be the first principal component to be learned. Let $X = [x_1, x_2, ..., x_N] \in \mathbb{R}^{D \times N}$, with $x_i = [x_{i1}, x_{i2}, ..., x_{ih}]^T \in \mathbb{R}^{h \times w}(i = 1,2,...,N)$. Note, $x_{ij} \in \mathbb{R}^{1 \times w}$ The objective of PCA-L1 is to maximize the L1-norm variance in feature space as follows:

$$f(w) = \|x_i w\|_1 = \sum_{i=1}^{N}\sum_{j=1}^{h} |x_{ij} w|$$

$$\text{subject to } \|w\|_2 = 1, \quad (5)$$

The polarity parameter $p_{ij}$ can be computed as:

$$p_{ij} = \begin{cases} 1, & \text{when } x_{ij}w \geq 0 \\ -1, & \text{when } x_{ij}w < 0. \end{cases} \quad (6)$$

The process of maximization is achieved by Algorithm 2.

| **Algorithm 2**: L1-2DPCA method |
|---|

Input:
- training set: $X = [x_1, x_2, ..., x_N] \in \mathbb{R}^{D \times N}$

Output:
- filters $w^*$

1: Set $w(0) = 0$ and $t = 0$

2: For all $i \in \{1, 2, ..., N\}$ and $j \in \{1, 2, ..., h\}$, calculate $p_{ij}(t)$ by using Equation (6).

3: Let $t = t + 1$ and $w(t) = \sum_{i=1}^{N}\sum_{j=1}^{h} p_{ij}(t-1)x_{ij}$. Then we initialize $w(t) = w(t)/\|w(t)\|_2$

4: If $w(t) \neq w(t-1)$, go back to Step 2. Otherwise, set $w^* = w(t)$ and stop.

To compute $w_k^*(k > 1)$, we have to update the training data as:

$$x_{ij}^k = x_{ij}^{k-1} - x_{ij}^{k-1}(w_{k-1}^* w_{k-1}^{*T}). \quad (7)$$

At this point, we can find that the difference between L1-PCA and L1-2DPCA is that L1-PCA converts the image matrix into vectors, and L1-2DPCA directly uses each row in the original image matrix as a vector.

### III. L1-PCANET

In this section, we propose a new PCA-based deep learning network, L1-PCANet. To overcome the sensitivity to outliers in PCANet due to the use of L2-norm, we use the PCA-L1 rather than the PCA to learn the filters. L1-PCANet and PCANet [9] share the same network architecture which is shown in Fig.1.

Suppose there are $N$ training images $I_i(i = 1,2,...,N)$ of size $m \times n$, and we get $D = m \times n$ patches of size $k \times k$ around each pixel in $I_i$. Then we take all overlapping patches and map them into vectors

$$[x_{i,1}, x_{i,2}, ..., x_{i,mn}] \in \mathbb{R}^{k^2 \times mn}. \quad (8)$$

And we remove the patch mean from each patch and get

$$\bar{X} = [\bar{x}_{i,1}, \bar{x}_{i,2}, ..., \bar{x}_{i,mn}] \in \mathbb{R}^{k^2 \times mn}. \quad (9)$$

For all input images, we construct the same matrix and combine them into one matrix to obtain

$$X = [\bar{X}_1, \bar{X}_2, ..., \bar{X}_N] \in \mathbb{R}^{k^2 \times Nmn}. \quad (10)$$

Then, we use L1-PCA mentioned above to learn the filters in stage 1. The filter we want to find is $w \in \mathbb{R}^{k^2 \times 1}$. We take $X$ as the input data of L1-PCA. Assuming that the number of filters in stage 1 is $L_1$, we can obtain the first stage filters $\{w_1^*, ..., w_{L_1}^*\}$ by repeatedly calling Algorithm 1. The L1-PCA filters of stage 1 are expressed as:

$$W_p^1 = mat_{k,k}(w_p^*) \in \mathbb{R}^{k \times k}, \quad (11)$$

where $p = 1,2,...,L_1$.

The output of stage 1 can expressed as:

$$O_i^p = I_i * W_p^1, i = 1,2,...,N, \quad (12)$$

where $*$ denote 2D convolution. We set the boundary of the input image to zero-padding to make sure that $O_i^p$ is of the same size as $I_i$. We can get the filters of second and subsequent layers by simply repeating the process of the first layer design. The pooling layer of L1-PCANet is almost the same as the pooling layer of L1-2D$^2$PCANet.

### IV. L1-2D$^2$PCANET

In this section, we generalize L1-PCANet to L1-2D$^2$PCANet. L1-2D$^2$PCANet and 2DPCANet [10] share the same network as shown in Fig.2.

## A. The first L1-2D²PCANet stage

Let all the assumptions be the same as in Section III. We get all the overlapping patches

$$x_{i,j} \in \mathbb{R}^{k \times k}, j = 1,2, \ldots, mn, \quad (13)$$

and subtract the patch mean from each of them and we form a matrix

$$\bar{X}_{x,i} = [\bar{x}_{i,1}, \bar{x}_{i,2}, \ldots, \bar{x}_{i,mn}] \in \mathbb{R}^{k \times kmn}. \quad (14)$$

And we use the transpose of $x_{i,j}$ to form matrix

$$\bar{X}_{y,i} = [\bar{x}_{i,1}{}^T, \bar{x}_{i,2}{}^T, \ldots, \bar{x}_{i,mn}{}^T] \in \mathbb{R}^{k \times kmn}. \quad (15)$$

For all input images, we construct the matrix by the same way and put them into one matrix, we can obtain

$$X_x = [\bar{X}_{x,1}, \bar{X}_{x,2}, \ldots, \bar{X}_{x,N}] \in \mathbb{R}^{k \times Nkmn}, \quad (16)$$

$$X_y = [\bar{X}_{y,1}, \bar{X}_{y,2}, \ldots, \bar{X}_{y,N}] \in \mathbb{R}^{k \times Nkmn}. \quad (17)$$

Then, we use L1-2DPCA mentioned above to learn the filters in Stage 1. We want to obtain filters $w^*_{x,p} \in \mathbb{R}^{k \times 1}$ and $w^*_{y,p} \in \mathbb{R}^{k \times 1}$, where $p = 1,2, \ldots, L_1$. $X_x$ and $X_y$ are the input data for 2DPCA-L1. Assuming that the number of filters in stage 1 is $L_1$, the first stage filters $\{w^*_{x,1}, \ldots, w^*_{x,L_1}\}$ and $\{w^*_{y,1}, \ldots, w^*_{y,L_1}\}$ are obtained by repeatedly calling Algorithm 2.

The filters we need in Stage 1 can finally be expressed as:

$$W^1_p = w^*_{x,p} \times w^*_{y,p}{}^T \in \mathbb{R}^{k \times k}. \quad (18)$$

The output of Stage 1 will be

$$O^p_i = I_i * W^1_p, i = 1,2, \ldots, N. \quad (19)$$

## B. The second L1-2D²PCANet stage

Like in the first stage, we can start with the overlapping patches of $O^p_i$ and remove the patch mean from each patch. Then we form

$$Y^p_{x,i} = [\bar{y}_{i,p,1}, \ldots, \bar{y}_{i,p,mn}] \in \mathbb{R}^{k \times kmn}. \quad (20)$$

$$Y^p_{y,i} = [\bar{y}_{i,p,1}{}^T, \ldots, \bar{y}_{i,p,mn}{}^T] \in \mathbb{R}^{k \times kmn}. \quad (21)$$

Further, we define the matrix that collects all the patches without the patch mean of the kth output $O^k_i$ being removed as:

$$Y^p_x = [Y^m_{x,1}, Y^m_{x,2}, \ldots, Y^m_{x,N}] \in \mathbb{R}^{k \times Nkmn}. \quad (22)$$

$$Y^p_y = [Y^p_{y,1}, Y^p_{y,2}, \ldots, Y^p_{y,N}] \in \mathbb{R}^{k \times Nkmn}. \quad (23)$$

Finally the input of the second stage is obtained by concatenating $Y^p_x$ and $Y^p_y$ for all $L_1$ filters

$$Y_x = [Y^1_x, Y^2_x, \ldots, Y^{L_1}_x] \in \mathbb{R}^{k \times L_1 Nkmn}. \quad (24)$$

$$Y_y = [Y^1_y, Y^2_y, \ldots, Y^{L_1}_y] \in \mathbb{R}^{k \times L_1 Nkmn}. \quad (25)$$

We take $Y_x$ and $Y_y$ as the input data of L1-2DPCA. Assuming that the number of filters in stage 2 is $L_2$, we design the second stage filters $\{w^*_{x,1}, \ldots, w^*_{x,L_2}\}$ and $\{w^*_{y,1}, \ldots, w^*_{y,L_2}\}$ by repeatedly calling Algorithm 2. The L1-2DPCA filters of Stage 2 are expressed as:

$$W^2_q = w^*_{x,q} \times w^*_{y,q}{}^T \in \mathbb{R}^{k \times k}, \quad (26)$$

where $q = 1,2, \ldots, L_2$.

Therefore, we have $L_2$ outputs for each output $O^p_i$ of Stage 1

$$B^q_i = \{O^p_i * W^2_q\}, l = 1,2, \ldots, L_2. \quad (27)$$

Note, the number of outputs of Stage 2 is $L_1 L_2$.

## C. The pooling stage

First, we use a Heaviside-like step function to binarize the output of Stage 2. The function $H(\cdot)$ can be expressed as:

$$H(x) = \begin{cases} 0, x < 0 \\ 1, x \geq 0. \end{cases} \quad (28)$$

Each pixel is encoded by the following function

$$T^m_i = \sum_l^{L_2} 2^{l-1} H(B^q_i), \quad (29)$$

where $T^m_i$ is an integer of range $[0, 2^{L_2}-1]$.

Second, we divide $T^m_i$ into $B$ blocks. Then we make a histogram of all blocks of $T^m_i$ with $2^{L_2}$ values, and concatenate all the histogram of $B$ blocks into one vector hist($T^m_i$). In this way, we obtain $L_1$ histograms and we put them into a vector

$$f_i = [\text{hist}(T^1_i), \ldots, \text{hist}(T^{L_2}_i)] \in \mathbb{R}^{2^{L_2} L_1 B \times 1}. \quad (30)$$

Using the 2DPCA-L1 model described above, we can transform an input image into a feature vector as the output of L1-2D²PCANet.

## V. EXPERIMENTS AND ANALYSIS

In this section, we evaluate the performance of L1-PCANet and L1-2D²PCANet with PCANet and 2DPCANet as baselines on YALE, AR, Extended Yale B and FRRET databases which are shown in Figure 3. The parameters are set as $k = 5$, $B = 8$, $L_1 = L_2 = 4$. SVM [22] implementation from the libsvm is used as the classifier with default settings. We repeat some experiments ten times and calculate the average recognition accuracy and root mean square error(RMSE).

### A. Extended Yale B

Extended Yale B consists of 2414 images of 38 individuals captured with different lighting conditions. These pictures are pre-processed to have the same size $48 \times 42$ and alignment.

In Experiment 1, we compare L1-PCANet and L1-2D$^2$PCANet with PCANet and 2DPCANet. We randomly select i(=2,3,4,5,6,7) images per individual for training and use the rest for testing. We also test a two-layer CNN for comparison which is trained on the test images for 2000 epochs. The architecture of CNN is the same as [10]. The results are shown in TABLE I.

In Experiment 2, to evaluate the robustness of L1-PCANet and L1-2D$^2$PCANet to outliers, we randomly add block-wise noise to the test images to generate test images with outliers. Within each block, the pixel value is randomly set to be 0 or 255. These blocks occupy 10%, 20%, 30%, and 50% of the images and they are added to the random position of the image, respectively which can be seen in see Figure 4. The results are shown in TABLE II.

In Experiment 3, we examine the impact of the block size B for L1-2D$^2$PCANet. The block size changes from $2 \times 2$ to $8 \times 8$. The rest parameters are the same as Experiment 1. The results are shown in Figure 5.

*B. AR*

AR face database contains 2600 color images corresponding to 100 people's faces (50 men and 50 women). It has two session data from two different days and each person in each session has 13 images including 7 images with only illumination and expression change and 3 images wearing sunglasses and 3 images wearing scarf. Images show frontal faces with different facial expressions, illumination conditions, and occlusions (sunglasses and scarf). These pictures are pre-processed to have the same size $40 \times 30$.

In Experiment 4, in order to investigate the impact of the choice of training images, we divide the experiment into four groups. (1) In group 1, we randomly select 5 images with only illumination and expression change from session 1 per individual as training images; (2) In group 2, we randomly select 4 images with only illumination and expression change and 1 image wearing sunglasses from session 1 per individual as training images; (3) In group 3, we randomly select 4 images with only illumination and expression change and 1 image wearing scarf from session 1 per individual as training images. The remaining images are test samples; (4) In group 4, we randomly select 3 images with only illumination and expression change, 1 image wearing sunglasses and 1image wearing scarf from session 1 per individual as training images. The remaining images in session 1 and all images in session 2 are used as test images. We manually select five images from session 1 as the gallery images and keep gallery images of each group the same. The results are shown in TABLE III.

In order to investigate the impact of the choice of gallery images, Experiment 5 is the same as Experiment 4 except that the gallery images and the training images are exchanged. We use the remaining images in session 1 and all images in session 2 as test samples. The results are shown in TABLE IV.

*C. FERET*

This database contains a total of 11338 facial images. They were collected by photographing 994 subjects at various facial angles. We gathered a subset from FERET which is composed by 1400 images recording of 200 individuals, with each 7 images exhibit large variations in facial expression, facial angle, and illumination. These pictures are pre-processed to have the same size $40 \times 40$ and alignment.

In Experiment 6, we divide the experiment into 7 groups. The training images of each group consist of 200 images from the subset with different facial angle, expression and illumination. We use the remaining images in the subset as test images. The parameters are set as $k = 5$, $B = 10$, $L_1 = L_2 = 4$. The results are shown in TABLE V.

*D. YALE*

YALE consists of 15 individuals and 11 images for each individual which shows varying facial expressions and configurations. These pictures are pre-processed to have the same size $32 \times 32$.

In Experiment 7, we randomly select i(=2,3,4,5,6,7) images per individual for training and use the rest for testing. The parameters are set as $k = 5$, $B = 4$, $L_1 = L_2 = 4$. The results are shown in TABLE VI.

*E. LFW-a*

LFW-a is a version of LFW after alignment with deep funneling. We gathered the individuals including more than 9 images and then form a dataset with 158 individuals from LFW-a.

In Experiment 8, we randomly choose i(=3,4,5,6,7) images per individual for gallery images and keep training images of each group the same. The results are shown in TABLE VII.

*F. Results and Analysis*

Table I shows the result of Experiment 1 on Extended Yale B, Table III shows the result of Experiment 4 on AR, Table V shows the result of Experiment 6 on FERET, Table VI shows the result on Yale, and Table VII shows the result on LFW-a. In these experiments, we changed the training images by random selection. From the results, we can see that the L1-2D$^2$PCANet outperforms PCANet, 2DPCANet and L1-PCANet in terms of recognition accuracy and RMSE, because we introduce L1-norm into the network. The two L1-norm-based networks we proposed are far superior to the traditional L2-norm-based networks in terms of RMSE, which means the proposed networks are insensitive to changes in training images. That is, the accuracy of the traditional L2-norm-based networks largely depends on the choice of training images while the L1-norm-based networks we proposed can achieve better and stable accuracy under any training images. A possible explanation of this phenomenon is as follows. In fact, the expression, posture, illumination condition and occlusion in the images can be regarded as interference or noise in face recognition. This noise degrades L2-norm-based networks much more than it degrades L1-norm-based networks. Therefore, the proposed networks exhibit its superiority when the training images contain some changes in expression, posture, illumination condition and occlusion.

Table II shows the result of Experiment 2 on Extended Yale B. In this experiment, we randomly add block-wise noise to the test images. From the results, we can see that as the block-wise noise increasing from 10% of the image size to 50%, the performance of PCANet, 2DPCANet, and L1-PCANet drops rapidly while L1-$2D^2$PCANet still has good performance. Therefore, it can be considered that L1-$2D^2$PCANet has better robustness against outlier and noise than other three networks.

We also investigate the impact of the choice of gallery images on AR; see Table IV. From the horizontal comparison of Table VI, the more categories the gallery contains, the higher the accuracy is.

Figure 5 shows the result of Experiment 3 on Extended Yale B. When the block is small, the local information cannot be contained perfectly, and it may get more noise when the block is too big.

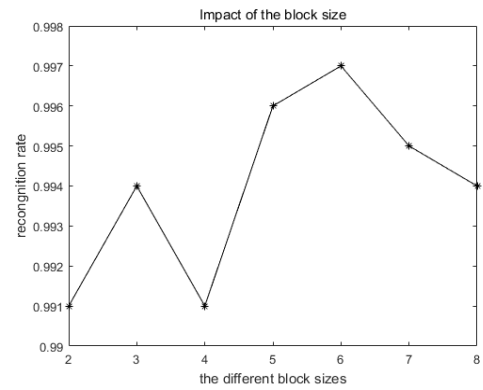

Fig.5: Impact of the block size.

TABLE I. Experiment 1 on Extended Yale B [19].

|  | 2 | 3 | 4 | 5 | 6 | 7 |
|---|---|---|---|---|---|---|
| CNN |  |  | 80.56 |  |  |  |
| PCANet | 83.41 ±5.31 | 84.51 ±5.70 | 84.42 ±5.37 | 82.48 ±7.18 | 84.06 ±6.22 | 89.56 ±5.48 |
| 2DPCANet | 97.48 ±1.03 | 97.34 ±1.81 | 97.01 ±1.64 | 96.71 ±2.48 | 95.16 ±2.93 | 97.22 ±2.02 |
| L1 − PCANet | 97.88 ±0.22 | 97.98 ±0.22 | 97.88 ±0.18 | 97.86 ±0.17 | 97.94 ±0.19 | 97.90 ±0.16 |
| L1 − $2D^2$PCANet | 99.67 ±0.09 | 99.71 ±0.07 | 99.73 ±0.09 | 99.73 ±0.06 | 99.75 ±0.06 | 99.77 ±0.07 |

TABLE II. Experiment 2 on Extended Yale B [19].

|  | 10% | 20% | 30% | 50% |
|---|---|---|---|---|
| PCANet | 92.68 ±0.42 | 88.51 ±0.40 | 74.63 ±0.48 | 44.10 ±0.76 |
| 2DPCANet | 94.26 ±0.25 | 88.71 ±0.57 | 79.54 ±0.89 | 55.34 ±0.70 |
| L1 − PCANet | 94.34 ±0.40 | 91.50 ±0.51 | 83.58 ±0.60 | 65.01 ±0.61 |
| L1 − $2D^2$PCANet | 99.00 ±0.15 | 98.28 ±0.18 | 95.73 ±0.20 | 84.01 ±0.74 |

TABLE III. Experiment 4 on AR [18].

|  | No occlusion | Sunglass | Scarf | Sunglass and Scarf |
|---|---|---|---|---|
| PCANet | 78.63 ±3.09 | 78.74 ±4.84 | 79.23 ±4.47 | 80.40 ±4.10 |
| 2DPCANet | 82.94 ±4.31 | 83.85 ±4.48 | 82.21 ±2.97 | 83.44 ±4.27 |
| L1 − PCANet | 87.09 ±0.50 | 86.73 ±0.31 | 87.33 ±0.12 | 86.46 ±0.22 |
| L1 − $2D^2$PCANet | 89.26 ±0.37 | 88.59 ±0.27 | 88.85 ±0.28 | 88.52 ±0.19 |

TABLE IV. Experiment 5 on AR [18].

|  | No occlusion | Sunglass | Scarf | Sunglass and Scarf |
|---|---|---|---|---|
| PCANet | 66.71 ±0.87 | 69.62 ±0.69 | 69.59 ±0.69 | 72.66 ±0.70 |
| 2DPCANet | 69.24 ±0.70 | 74.78 ±0.70 | 72.14 ±0.99 | 75.51 ±0.61 |
| L1 − PCANet | 68.56 ±0.65 | 75.23 ±0.60 | 72.35 ±0.77 | 79.34 ±0.71 |
| L1 − $2D^2$PCANet | 77.08 ±0.64 | 81.10 ±0.37 | 78.34 ±0.61 | 84.17 ±0.75 |

TABLE V. Experiment 6 on FERET [21].

|  | 1 | 2 | 3 | 4 | 5 | 6 | 7 | Avg. | RMSE |
|---|---|---|---|---|---|---|---|---|---|
| PCANet | 75.83 | 76.83 | 76.17 | 68.00 | 73.67 | 69.83 | 79.11 | 74.21 | 3.69 |
| 2DPCANet | 73.17 | 76.17 | 76.17 | 73.67 | 78.33 | 73.50 | 74.00 | 75.00 | 1.78 |
| L1 − PCANet | 82.83 | 82.17 | 82.00 | 82.50 | 85.00 | 82.50 | 81.83 | 82.69 | 0.99 |
| L1 − $2D^2$PCANet | 86.00 | 84.83 | 85.50 | 86.50 | 87.33 | 86.83 | 86.83 | 86.26 | 0.81 |

TABLE VI. Experiment 7 on Yale [19].

| | 2 | 3 | 4 | 5 | 6 | 7 |
|---|---|---|---|---|---|---|
| PCANet | 86.33 ±1.87 | 86.75 ±2.37 | 87.50 ±1.58 | 87.25 ±2.12 | 87.25 ±2.14 | 87.29 ±2.22 |
| 2DPCANet | 91.33 ±2.80 | 91.78 ±1.94 | 90.44 ±2.59 | 90.67 ±2.34 | 90.87 ±2.90 | 91.93 ±2.13 |
| L1 − PCANet | 91.45 ±0.89 | 92.00 ±0.83 | 91.22 ±0.54 | 91.00 ±0.44 | 91.89 ±0.51 | 92.67 ±0.33 |
| L1 − 2D$^2$PCANet | 94.03 ±0.32 | 95.10 ±0.41 | 94.95 ±0.33 | 95.25 ±0.32 | 95.16 ±0.41 | 95.66 ±0.40 |

TABLE VII. Experiment 8 on LFW-a [20].

| | 3 | 4 | 5 | 6 | 7 |
|---|---|---|---|---|---|
| PCANet | 30.07 ±4.69 | 31.86 ±5.35 | 34.35 ±5.91 | 35.71 ±6.34 | 38.56 ±6.82 |
| 2DPCANet | 33.00 ±3.52 | 35.68 ±3.64 | 39.02 ±3.74 | 39.92 ±3.98 | 43.15 ±4.12 |
| L1 − PCANet | 34.14 ±0.39 | 36.27 ±0.29 | 39.08 ±0.57 | 40.25 ±0.77 | 44.26 ±0.81 |
| L1 − 2D$^2$PCANet | 39.35 ±0.29 | 42.20 ±0.46 | 45.91 ±0.34 | 46.99 ±0.42 | 50.12 ±0.47 |

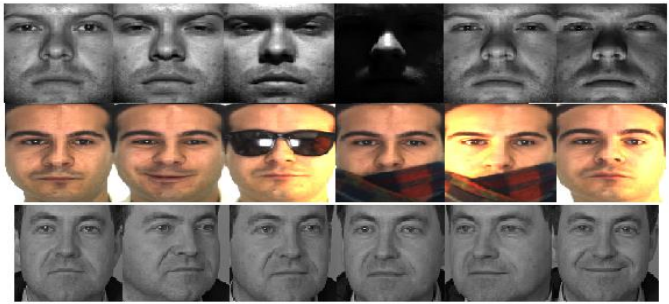

Fig.3: images in three datasets. Top line: Extended Yale B [19]. Middle line: AR [18]. Bottom line: FERET [21].

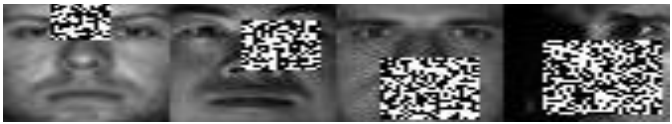

Fig.4: some generalized outlying face images of Extended Yale B [19].

## VI. CONCLUSION

In this paper, we have proposed a new deep learning network L1-2D$^2$PCANet, which is a simple but robust method. We use the L1-norm-based 2DPCA [17] instead of L2-norm-based 2DPCA [11] for the filter learning because of the advantages of L1-norm. It is more robust to outliers than L2-norm. By introducing L1-norm into 2DPCANet [10], we hope the network will inherit such advantages.

To verify the performance of L1-2D$^2$PCANet, we evaluate them on the facial datasets including AR, Extended Yale B, Yale and FRRET. The results show that L1-2D$^2$PCANet has three distinct advantages over traditional L2-norm-based networks: (1) Statistically, the accuracy of L1-2D$^2$PCANet is higher than that of other networks on all test datasets. (2) L1-2D$^2$PCANet has better robustness to changes in training images compared with the other networks. (3) Compared with the other networks, L1-2D$^2$PCANet has better robustness to noise and outliers. Therefore, L1-2D$^2$PCANet is an efficient and robust network for face recognition.

However, L1-2DPCA brings more computational load to the network, which increases the computational cost of L1-2D$^2$ PCANet. Despite this, the computational cost of L1-2D$^2$PCANet is far less than those traditional CNNs which are based on back propagation.

In the future work, we will work on the improving of L1-2DPCA algorithm to solve the problem of the computational cost of L1-2D$^2$PCANet.